\documentclass[a4paper,twoside]{article}

\usepackage{epsfig}
\usepackage{subcaption}
\usepackage{calc}
\usepackage{amssymb}
\usepackage{amstext}
\usepackage{amsmath}
\usepackage{amsthm}
\usepackage{multicol}
\usepackage{pslatex}
\usepackage{apalike}
\usepackage{booktabs}
\usepackage{multirow}
\usepackage{SCITEPRESS}     

\usepackage{todonotes}

\begin{document}

\title{Solving Large Steiner Tree Problems in Graphs for Cost-Efficient Fiber-To-The-Home Network Expansion}

\author{\authorname{
Tobias Müller\sup{1},
Kyrill Schmid\sup{1},
Daniëlle Schuman\sup{1},
Thomas Gabor\sup{1},
Markus Friedrich\sup{1} and
Marc Geitz\sup{2}}

\affiliation{\sup{1}Mobile and Distributed Systems Group, LMU Munich, Germany}
\affiliation{\sup{2}Telekom Innovation Laboratories, Deutsche Telekom AG, Bonn, Germany}
\email{\{tobias.mueller1, d.schuman\}@campus.lmu.de, \{kyrill.schmid, thomas.gabor, markus.friedrich\}@ifi.lmu.de, marc.geitz@telekom.de}
}


\abstract{
The expansion of Fiber-To-The-Home (FTTH) networks creates high costs due to expensive excavation procedures. Optimizing the planning process and minimizing the cost of the earth excavation work therefore lead to large savings. Mathematically, the FTTH network problem can be described as a minimum Steiner Tree problem. Even though the Steiner Tree problem has already been investigated intensively in the last decades, it might be further optimized with the help of new computing paradigms and emerging approaches. This work studies upcoming technologies, such as Quantum Annealing, Simulated Annealing and nature-inspired methods like Evolutionary Algorithms or slime-mold-based optimization. Additionally, we investigate partitioning and simplifying methods. Evaluated on several real-life problem instances, we could outperform a traditional, widely-used baseline (NetworkX Approximate Solver \cite{networkx}) on most of the domains. Prior partitioning of the initial graph and the slime-mold-based approach were especially valuable for a cost-efficient approximation. Quantum Annealing seems promising, but was limited by the number of available qubits.
}

\onecolumn \maketitle \normalsize \setcounter{footnote}{0} \vfill

\section{\uppercase{Introduction}}
\label{sec:introduction}
\begin{figure*}[hpbt]
 \centering
  \subfloat[Grid and House Connections]{
   \label{fig:cadastral:a}
   \includegraphics[width=0.32\linewidth]{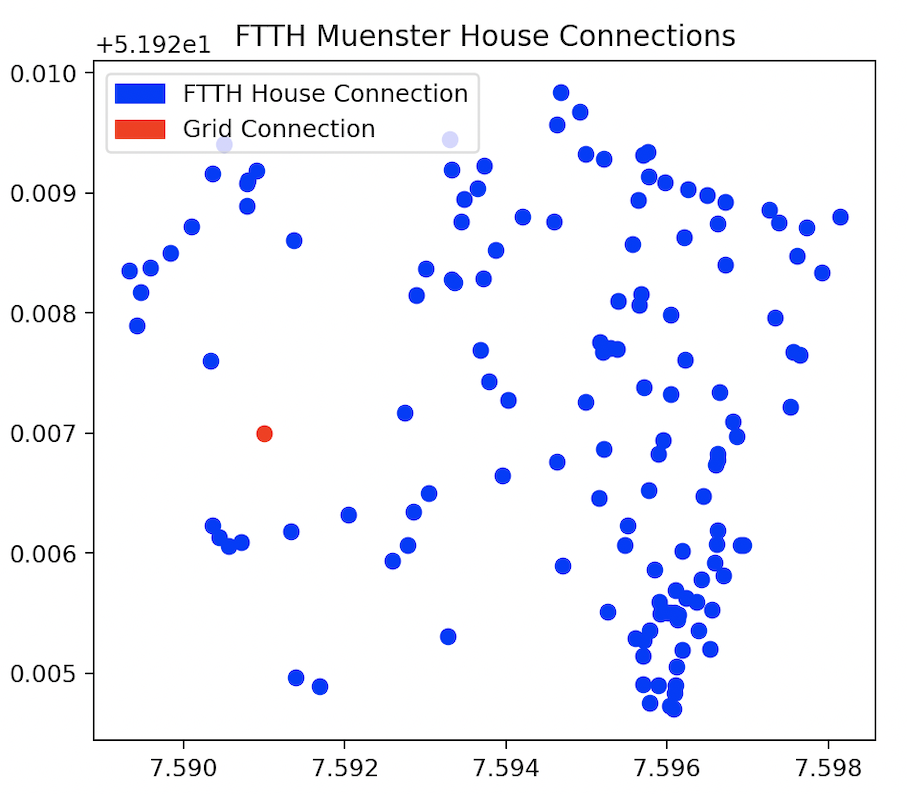}}
  \subfloat[Connections and Streets]{
   \label{fig:cadastral:b}
   \includegraphics[width=0.33\linewidth]{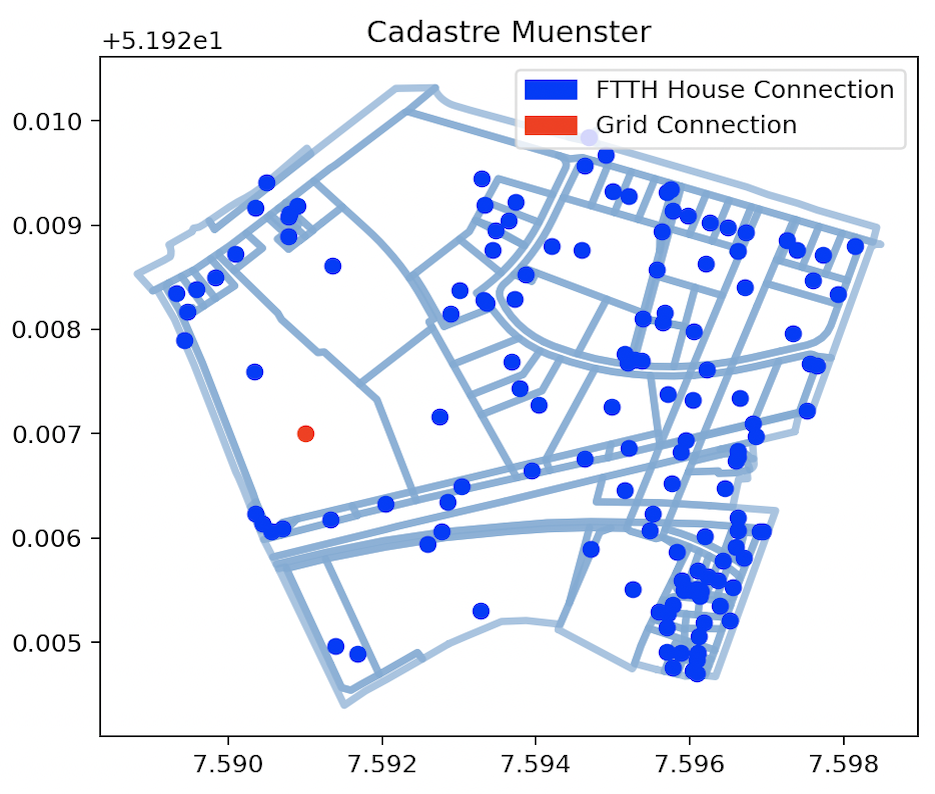}}
  \subfloat[Ditch Cost Map]{
   \label{fig:cadastral:c}
   \includegraphics[width=0.32\linewidth]{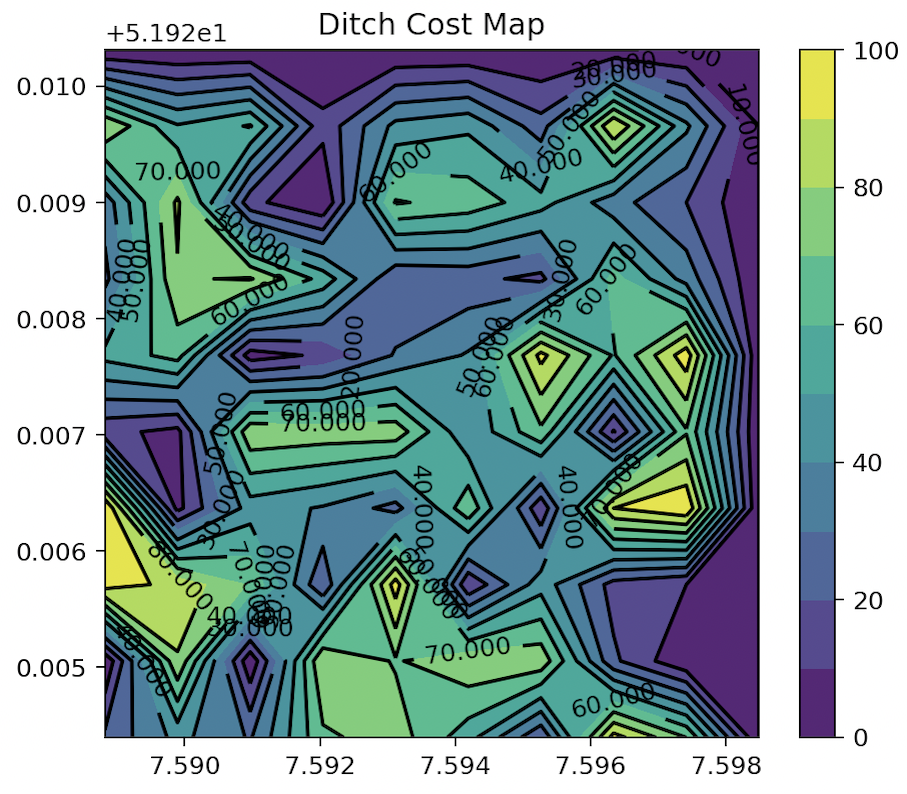}}
 \caption[]{Example of a cadastral excerpt (Figure \ref{fig:cadastral:a}) with mapped FTTH house connections (blue) and access nodes (red). Figure \ref{fig:cadastral:b} additionally displays street courses. The ditch cost map (Figure \ref{fig:cadastral:c}) provides the excavation costs per meter.}
 \label{fig:cadastral-data}
\end{figure*}

Internet traffic is constantly increasing over time due to growing digitization and the increasing use of bandwidth intensive applications. Internet consumers, be it large industry, small enterprises or private households require glass fiber (FTTH) connection to meet the increasing demand. The main cost driver of fiber roll-out for land lines is the earth excavation cost. Optimizing the planning process and finding better networks can reduce needed excavation, which leads to large savings. Figure \ref{fig:cadastral-data} visualizes an exemplary cadastral excerpt with house connections and access nodes for which an optimal network needs to be found with respect to the ditch cost map on the right.

Finding a cost-minimal solution for connecting multiple FTTH households can be described as a Steiner Tree Problem \cite{promel2012steiner}. Calculating minimum Steiner Trees is a well-known problem and has already been investigated in a variety of network approximation use cases \cite{gupta11}. However, the emergence of new computing paradigms and novel approaches opens up new possibilities to tackle the Steiner Tree problem. 

This work has the goal to analyse emerging solution methods such as nature inspired algorithms and Quantum Annealing technologies to find optimal network configurations for a given problem instance. Since we want to solve real-life problems, we focused on developing a practical, easy-to-use application. Therefore, we implemented a Python-based demonstrator, where cadaster data can be easily imported and transformed into a corresponding Steiner Tree problem. We implemented various methods to solve, simplify and partition the resulting graph. 

We chose nature-inspired solution methods, such as an Evolutionary Algorithm (EA) \cite{rosenberg21} or a Physarum solver based on the behavior of the Physarum polycephalum slime mold \cite{sun19}. We also included a Simulated Annealing (SA) algorithm which finds solutions by building a long Markov Chain \cite{grimwood94} and a Quantum Annealing (QA) solver based on a Quadratic Unconstrained Binary Optimization (QUBO) formulation of the Steiner Tree problem \cite{Lucas14}. 

These different solution methods have been compared and evaluated against a commonly known baseline (NetworkX approximated Steiner Tree solver \footnote{https://networkx.org/documentation/}).
For larger instances the solvers mentioned above may become intractable due to the exponential increase in complexity. Therefore, we also provide methods to simplify or partition a given problem and run solvers on the simplified (respectively partitioned) graph. After a simplifier or partitioner has been applied, the problem can be passed again into a solver method, thereby decreasing the duration of the optimization significantly.

Summarized, our contributions are two-fold:

\begin{itemize}
    \item We provide a wide range of state-of-the-art algorithms for solving, simplifying and partitioning large Steiner Tree problems in graphs. These methods are comprised in a demonstrator.
    \item We provide a thorough evaluation of these methods on real-world problems and their impacts on resulting network costs and runtime.
\end{itemize}
\section{\uppercase{Preliminaries}}
\label{sec:preliminaries}
\subsection{The Steiner Tree Problem}
Formally, the problem of finding a cost-minimal network to connect $N$ FTTH house connections with $M$ access nodes can be described as a Steiner Tree Problem (STP) \cite{promel2012steiner}.
The formulation of the STP is defined as follows. Let $G$ be an undirected weighted graph $G=(V,E)$, with $V$ as the set of nodes and $E$ as the set of edges. The set of nodes $V$ defines a disjoint set of \textit{terminal nodes} $T \subseteq V$ and potential \textit{Steiner nodes} $S=V \setminus T$. Finally, a cost function $c \colon E \rightarrow \mathbb{R^{+}}$ assigns a non-negative value to each edge.

We aim to find a cost-minimal subgraph $G_T$ that spans all terminals, such that $S \subseteq V_T \subseteq V, E_T \subseteq E$ and $min(\sum_{e \in E_T} c(e))$. The minimum STP is a $\mathcal{NP}$-hard combinatorial optimization problem \cite{leitner2014partition,hwang92}. 

\begin{figure}[hpbt]
 \centering
  \subfloat[Terminal nodes]{
   \label{fig:STP:a} 
   \includegraphics[width=0.3\linewidth]{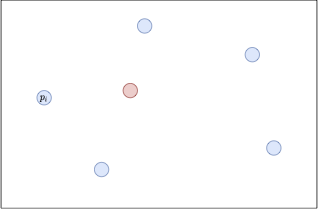}}
   \hspace{0.004\linewidth}
  \subfloat[All nodes]{
   \label{fig:STP:b} 
   \includegraphics[width=0.3\linewidth]{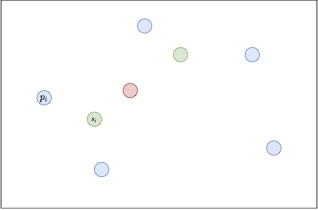}}
   \hspace{0.004\linewidth}
  \subfloat[Steiner Tree]{
   \label{fig:STP:c} 
   \includegraphics[width=0.3\linewidth]{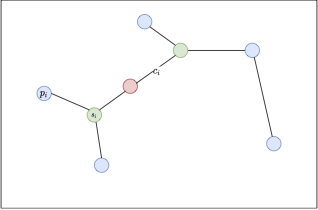}}
 \caption[]{An STP for a set of terminal nodes (red and blue) and non-terminal nodes (green)}
 \label{fig:STP} 
\end{figure}

Figure \ref{fig:STP} shows how a Steiner Tree is build upon a set of terminal nodes (blue and red, Figure \ref{fig:STP:a}) by including non-terminal nodes (green, Figure \ref{fig:STP:b}) and finding a set of edges connecting them for given edge costs (Figure \ref{fig:STP:c}).

\subsection{Demonstrator}
In order to develop, compare and visualize algorithms for the STP during the course of the project a Python based demonstrator has been developed. Problem in- stances from various sources (e.g. pixel-based images or Keyhole Markup Language (KML) cadaster data) are directly imported and transformed into a corre- sponding STP. For image based imports each pixel in the image is interpreted as a node, its color indi- cates the node type and if the node type is a way- point (non-terminal), its brightness value is used for its edge weights. Green pixels are interpreted as endpoints, whereas yellow pixels are considered distrib- utor nodes. All methods for simplifying, partitioning and solving provided by the demonstrator will be de- scribed in the following chapter.
\section{\uppercase{Approach}}
\label{sec:approach}
We approached the minimum STP with a high variety of state-of-the-art methods, ranging from population-based EAs and heuristic approximation techniques like Simulated Annealing to Quantum-Annealing technologies and slime-mold-inspired optimization.
Additionally, we implemented graph partitioning algorithms to tackle large graphs in a divide-and-conquer-like approach to reduce computational costs and approximation errors.
Similar to partitioning, we tried to reduce complexity by simplifying large graphs before running solvers. The following chapter will present all used methods in more detail.

\subsection{Simplifier}
Simplifiers are used to generate simplified graphs in order to make the application of solvers tractable.

\paragraph{Triangle Simplifier}
The triangle simplifier uses an EA to solve a combinatorial optimization problem which leads to a simplification of the input graph.
It is based on triangulation approaches for image compression \cite{lehner2008video} and tries to find a selection of non-terminals that, together with all terminal nodes, forms a Delaunay triangulation which approximates the input graph as good as possible.
The difference between node weights of the input graph and interpolated node weights from the simplified graph (at the node positions of the input graph) should be as small as possible. 
Thus, an individual $x$ represents a selection of non-terminals which are ranked using the fitness function:
\begin{equation}\label{eq:tri_obj_func}
    f(x) = -(\alpha \cdot \text{error} + \beta \cdot \text{size}(x)),
\end{equation}
where $\text{size}(x)$ is a normalized size metric that penalizes large non-terminal node selections. The error metric $\text{error}(\cdot)$ is defined as:
\begin{equation}
    \text{error} = \frac{1}{|N|} \sum_{n \in N}|w(p_n)-w_n|,
\end{equation}
where $N$ is the set of nodes of the input graph ($p_n$ is the position of node $n$, $w_n$ its weight) and $w(\cdot)$ is the weight function  that returns a weight value for a position by interpolating the weight value based on the weight values of the three corner points of the triangle the query position is located in. 
Please note that the error value for each individual is normalized based on the error values of all individuals in the current population (like the size penalty term in Equation \ref{eq:tri_obj_func}) in order to make it easier to find suitable values for parameters $\alpha$ and $\beta$. 
\\
It is possible to select a maximum number of non-terminals the simplified graph should contain which opens up the possibility to compare this approach to other simplification approaches. 

\paragraph{Growing Neural Gas Simplifier}
Growing Neural Gas (GNG) \cite{fritzke1994growing} is a method to learn important topological relations for a given unlabelled input data set.
In our case, the input data set is represented by the cost function $c$.
The algorithm iteratively builds a network structure comprising a set of nodes $A$ with an associated reference vector (such as position) and a set of unweighted edges $N$. The idea is to start with a comparatively small network and increase the nodes within that network by evaluating specific statistical measures. The high-level steps of the algorithm are as follows \cite{fritzke1994growing}: 
\begin{enumerate}
    \item Initialize the network (e.g. with two nodes).
    \item Generate an input signal $x$, i.e. a sample point in the input data $c$.
    \item Find the nearest and second nearest units $s_{1}, \, s_{2}$ in the network.
    \item Increment the age of all edges emanating from $s_{1}$.
    \item Add the squared distance between $x$ and $s_1$ to the counter variable $error(s_1)$.
    \item Move $s_1$ and its neighbors towards $x$. 
    \item Set the age of the edge between $s_1$ and $s_2$ to $0$.
    \item Remove edges older than a predefined threshold.
    \item Insert new units in places with the largest error.
    \item Decrease all error variables by multiplying them with a constant.
\end{enumerate}
One of the key advantages besides its algorithmic simplicity is that it has relatively few critical parameters.

\paragraph{Iso-level adaptation}
The standard GNG approach as described above develops a rather homogeneous network structure over the defined input space. In order to develop a network structure that can adapt with different numbers of nodes to different iso-cost levels in the ditch-cost matrix, the iso-level adaptation flag can be set. If set, the network samples nodes according to the different probabilities statically associated to different iso-levels. In principle, the so created network will develop more nodes in regions with high ditching costs so as to provide more planning nodes for the solver. In contrast, regions with low ditching costs will yield less nodes in the network which require less resources during the solving procedure.

\subsection{Partitioner}
By partitioning the problem instance represented by input graph $G$ and cost function $c$, the complexity and computational cost can be reduced. Firstly, $G$ is partitioned into several smaller STPs which are solved separately. 
Then all solutions are merged, resulting in a single resulting Steiner Tree. 
The following describes all partitioning and merging algorithms used in this work.

\paragraph{Greedy Modularity}
The Clauset-Newman-Moore Greedy Modularity maximization (GM) \cite{clauset04} is a clustering method based on community detection and works as follows:

\begin{enumerate}
    \item Every node belongs to a different community.
    \item The joined pair of communities which maximizes the modularity $M$ is merged, where $M$ is calculated for the whole graph.
    \item Step 2 is executed until a single community remains.
    \item The network partition with the highest modularity value is chosen. 
\end{enumerate}
Let $e_{ij}$ be the fraction of edges connecting vertices of group $i$ to group $j$, then the modularity $M$ is given by:

\begin{equation}
    M =  \sum_{i}(e_{ii}-(\sum_{j}e_{ij})^{2}),
\end{equation}
The higher the modularity, the higher the quality of the partition of the network, in the sense that there are many connections within communities and few between.

\paragraph{Spectral Clustering}
We chose to implement the Normalized Spectral Clustering algorithm (SC) according to Shi and Malik \cite{shi00} with the following clustering procedure:

\begin{enumerate}
    \item Get the normalized Laplace-Matrix $L_{rw}$ from input graph $G$.
    \item Compute the $k$ Eigenvectors corresponding to the $k$ largest Eigenvalues of $L_{rw}$.
    \item Cluster the columns of the $k$ Eigenvectors with $k$-Means Clustering.
\end{enumerate}
$L_{rw}$ is computed via $L_{rw} = D^{-1} * L$ with $L=D-W$ and $D$ as the Degree-Matrix and $W$ as Adjacency-Matrix of $G$.
The number of clusters $k$ can either be automatically derived or set manually. 
For the automatic determination, the following methods are used:

\textbf{Eigengaps.} The Eigengaps (jumps in the sequence of Eigenvalues) are analyzed based on Perturbation Theory and Spectral Graph Theory \cite{manor04}. 
The heuristic suggests to choose a value for $k$ which maximizes the Eigengaps. Our experiments revealed that the Eigengaps heuristic works best for graphs with evident clusters and is not optimal otherwise.

\textbf{Educated Guess.} Setting $k = \lceil \frac{m}{7} \rceil + 1$ for $m$ terminal nodes was found to return feasible solutions. However, this is not likely to be optimal.

\paragraph{Voronoi-based Partitioning}
For the Voronoi-based partitioning algorithm (Voronoi), we follow Leitner et al. (2014). 
First, the shortest path of every node $i$ in $G$ to all terminals $j$ is computed and then every node $i$ is assigned to its nearest terminal. By this, a Voronoi-diagram with corresponding Voronoi-regions is constructed.
Thereafter, the smallest Voronoi-region is merged with its nearest neighboring cluster if the overall number of nodes does not exceed a predetermined limit. 
This step is repeated until the target number of clusters $k$ has been reached \cite{leitner2014partition}.

\paragraph{Merging} \label{sec:Merging}
Two different procedures for merging the solutions of subgraphs are used. 
\textit{Exact} merging based on the Shortest Path Heuristic (SPH) or \textit{Center-of-Mass} merging.

For SPH, the shortest path of every node $i$ of subgraph $m$ of $G$ is calculated for every node $j$ of all other subgraphs of $G$. 
Node $i$ and $j$ with the shortest paths are connected. 
Since computing the Manhattan-Distance is faster than computing the shortest path, the merging procedure can be accelerated by preselecting subgraphs and nodes.
Hence, the Manhattan-Distance is computed between every combination of partitions whereas the shortest path is only determined for a specific number of nearest partitions. 
The same procedure is used for every combination of nodes.

As the name \textit{"Center-of-Mass"} suggests, the center of mass of each subgraph is derived. 
Subsequently, the shortest path between each center is calculated and the subgraphs with shortest lengths are merged.

\subsection{Solver}
\paragraph{NetworkX-Approximate Solver}
According to their documentation, the NetworkX approximate Steiner Tree solver works by computing the minimum spanning tree of the subgraph of the metric closure of the graph $G$. Edges are weighted by the shortest path distance between the nodes in $G$. The algorithm produces a result which is within a factor of $ (2 - (2 / t)) $ of the optimal Steiner tree where $t$ is the number of terminal nodes \cite{networkx}. We use the NetworkX-Approximate solver as the baseline.

\paragraph{Evolutionary Algorithm}\label{ch:ea_solver}
Evolutionary Algorithms (EAs) are nature-inspired and population-based meta-heuristics well-suited for solving combinatorial optimization problems like STP \cite{kapsalis1993solving}.
The optimization process starts with the initialization of a population of Steiner Trees consisting of randomly selected non-terminals and all terminals. Each individual (Steiner Tree) $x$ in the population $X$ is then ranked using a so-called fitness function 
\begin{equation}
    f(x) = -(\alpha \cdot \text{cost}(x) + \beta \cdot \text{size}(x)),
\end{equation}
where $\text{cost}(\cdot)$ is a function that returns a cost value $\in [0,1]$ for a Steiner Tree which is usually based on the edge weights of the input graph. $\text{size}(\cdot)$ penalizes tree size and uses a normalized size value based on all tree sizes in the population.
$\alpha$ and $\beta$ are user-defined weighting parameters.
\\
The best ranked Steiner Trees are selected as operands for the variation operators \textit{Recombination} and \textit{Mutation}. 
The \textit{Recombination} (or \textit{Crossover}) operator takes two parent trees as input and recombines them to two new trees. It works by replacing a randomly selected sub-tree from the first tree with the largest fitting sub-tree from the second tree (and vice versa). 
The \textit{Mutation} operator alters a single individual by either adding new non-terminals, removing selected terminals or creating a new sub-tree with existing terminals and non-terminals. Furthermore, with a certain user-controlled probability $\gamma$, a complete Steiner Tree is replaced by a newly, randomly created one.\\
The newly created individuals are part of the population of the next iteration, together with the $n$ best individuals of the current population. 
The process continues until a certain number of iterations has been reached.\\
The EA initializes the population partially with solutions of the baseline approach.
This way, the EA finds tiny pieces in the almost-optimal baseline solutions that can be improved and is able to achieve slightly better results (compared to baseline) in almost all tested problem instances.

\paragraph{Physarum}
Physarum Polycephalum is a non-intelligent slime mold with the ability to approximate shortest paths from its inoculation site to a source of nutrients \cite{adamatzky12,adamatzky14}. With multiple food sources, this cellular structure can be used to compute minimal Steiner Trees \cite{liu15}. In nature Physarum spreads itself throughout the environment and retracts itself from everywhere except the shortest route connecting the food sources by iteratively transporting a sort of fluid through its tubular structure. Inspired by this behavior, the Physarum solver by Y. Sun (2019) works as follows:
\begin{enumerate}
    \item Randomly select source and sink nodes from the set of terminals $T$.
    \item Calculate pressure $P_t$ for each terminal $t \in T$.
    \item Calculate flux $Q_{ij}$ for each edge $e_{ij}$. 
    \item Derive corresponding conductivities $D_{ij}$.
    \item Cut edge $e_{ij}$ using threshold $\epsilon$ for $Q_{ij}$.
\end{enumerate}
More specifically, the algorithm separates all terminal nodes into source respectively sink nodes and thus defines how the flux will flow through the network (from source to sink). Then, the pressure of each terminal $t$ is computed via
\begin{equation}
    P_t = \left\{\begin{array}{ll} \frac{-I_{0}}{n}, & \text{if} \:  t = \text{source} \\
             \frac{I_{0}}{m} & \text{if} \: t = \text{sink} \end{array}\right.,
\end{equation}
with $n$ source nodes, $m$ sink nodes and predefined initial pressure $I_0$. We set $I_0 = 1$. Using these pressure values, the flux $Q_{ij}$ of each edge $e_{ij}$ connecting nodes $i$ and $j$ can be computed by
\begin{equation}
    Q_{ij} = \frac{D_{ij}}{C_{ij}}*(P_{i}-P_{j}),
\end{equation}
where $C_{ij}$ is the cost for edge $e_{ij}$. 
$D_{ij}$ is the edge conductivity, which is initialized with $1$ if an edge exists between $i,j$ and $0$ otherwise. $D_{ij}$ is updated in each iteration. If $Q_{ij}$ falls below a user-defined threshold $\epsilon$ (in the evaluation: $\epsilon = 0.001$), the edge $e_{ij}$ is cut.
$D_{ij}$ is updated by
\begin{equation}
    D_{ij} = D_{ij} + \alpha|Q_{ij}| - \mu D_{ij},
\end{equation}
where $\alpha$ and $\mu$ are two positive constants (in the evaluation: $\alpha = 0.15$, $\mu = 1$). 
This process is repeated for $k$ iterations using $l$ initializations in total. 
To guarantee a coherent graph, the minimal spanning tree of the resulting graph is computed. Additionally, nodes which are not connected to a terminal and non-terminals with only one outgoing edge are cut off \cite{sun19}.

\paragraph{Quantum Annealing} \label{sec:QA}
Quantum Annealing (QA) is a restricted form of adiabatic quantum computation \cite{Vene18} that solves a specific group of optimization problems by exploiting quantum phenomena such as superposition, entanglement and quantum tunneling.
The corresponding cost functions to be minimized by the Quantum Annealer are formulated as Quadratic Unconstrained Binary Optimization Problems (QUBOs) \cite{Vene18}.
For the STP, such a QUBO-formulation exists \cite{Lucas14}, consisting of the following constraints from a high-level perspective:
 \begin{enumerate}
     \item The Steiner Tree has exactly one root.
     \item Each terminal node has a specified depth in the tree, as has each Steiner node contained in the tree.
     \item Each edge in the tree has a specified depth $i$. It connects two nodes of depths $i-1$ and $i$.
     \item The tree is connected: Apart from the root, every node has exactly one incoming edge from a node at lower depth.
     \item Summarized costs of all edges should be minimized.
 \end{enumerate}
The used QUBO-formulation of the STP needs $|V|( \lfloor |V| + 1 \rfloor + 4 + 2 |E|)/2 + |E|$ logical qubits for $|V|$ vertices and $|E|$ edges \cite{Lucas14}. According to this formulation, our smallest problem instance with $|V|=158$ and $|E|=458$ would require 85699 logical qubits.
Current Quantum Annealers are restricted to a maximum of about $5436$ physical qubits, which each have at most 15 connections to each other \cite{advantage}.
Thus, depending on problem size and structure, multiple physical qubits may be necessary to represent a single logical qubit, namely if it has more connections to other qubits \cite{Vene18}.
This restricts the set of problem instances solvable on QA hardware to only practically irrelevant instances.  
Therefore, two options remain:
\begin{itemize}
    \item Using a Quantum Annealing simulation software that does not impose these restrictions but is not able to exploit quantum effects (qbsolv).
    \item Using a quantum classical hybrid solver offered by an external cloud provider (Leap). \footnote{Leap Hybrid Solver Service, D-Wave Systems, https://www.dwavesys.com/take-leap}
\end{itemize}
These strategies neither guarantee solution optimality nor correctness. 
Thus, results can take the incorrect form of non-connected graphs.
Therefore, graph components are merged in a post-processing step after the annealing process if necessary.

\paragraph{Simulated Annealing}
The Simulated Annealing algorithm (SA), based on the Metropolis algorithm \cite{metropolis1953equation}, works by building a Monte Carlo Markov Chain (MCMC) to sample solution candidates from a desired (i.e.~cost minimal) target distribution. To build the chain, the algorithm works by performing the following steps to derive a new state $G_{t+1}$ (Steiner Tree) from a given state (Steiner Tree) $G_t = i$ \cite{madras2002lectures}:

\begin{enumerate}
    \item Choose a random "proposal" tree $Y \in \mathbf{V}$ according to probability transition matrix $Q$, i.e., $Pr(Y=j | G_t = i) = q_{ij}$.
    \item Define acceptance probability $\alpha = min\{1, \pi_Y / \pi_i \}$. 
    \item Accept $Y$ with probability $\alpha$, i.e., set $G_{t+1} \leftarrow Y$ with probability $\alpha$, and $G_{t+1} \leftarrow G_{t}$ with probability $1 - \alpha$.
\end{enumerate}
In order to build a Markov Chain for optimization problems it is necessary to define a target probability distribution that makes better solutions more likely. Such a target distribution is defined by the Gibbs distribution:

\begin{equation}
\pi^{\beta}(z) = \frac{e^{-\beta c(z)}}{C_{\beta}},
\end{equation}

where $C_{\beta}$ is the (unknown) normalizing constant. For $\beta \rightarrow \infty$ this distribution is concentrated on ground states, i.e. states with optimal configuration. For practical usage, it is important that the unknown normalizing constant cancels out, so the distribution can be easily computed at any time. 
\label{sec:evaluation} \section{\uppercase{Evaluation}}
In order to evaluate the described simplification, partitioning and solving procedures, several different problem instances were created.
Focus is thereby on resulting network cost (Steiner Tree size) and wall-clock times.

\begin{figure*}[hpbt]
   \subfloat[PI-$1$($1560$, $4294$)]{
    \label{fig:benchmark:1} 
    \fbox{\includegraphics[width=0.17\linewidth]{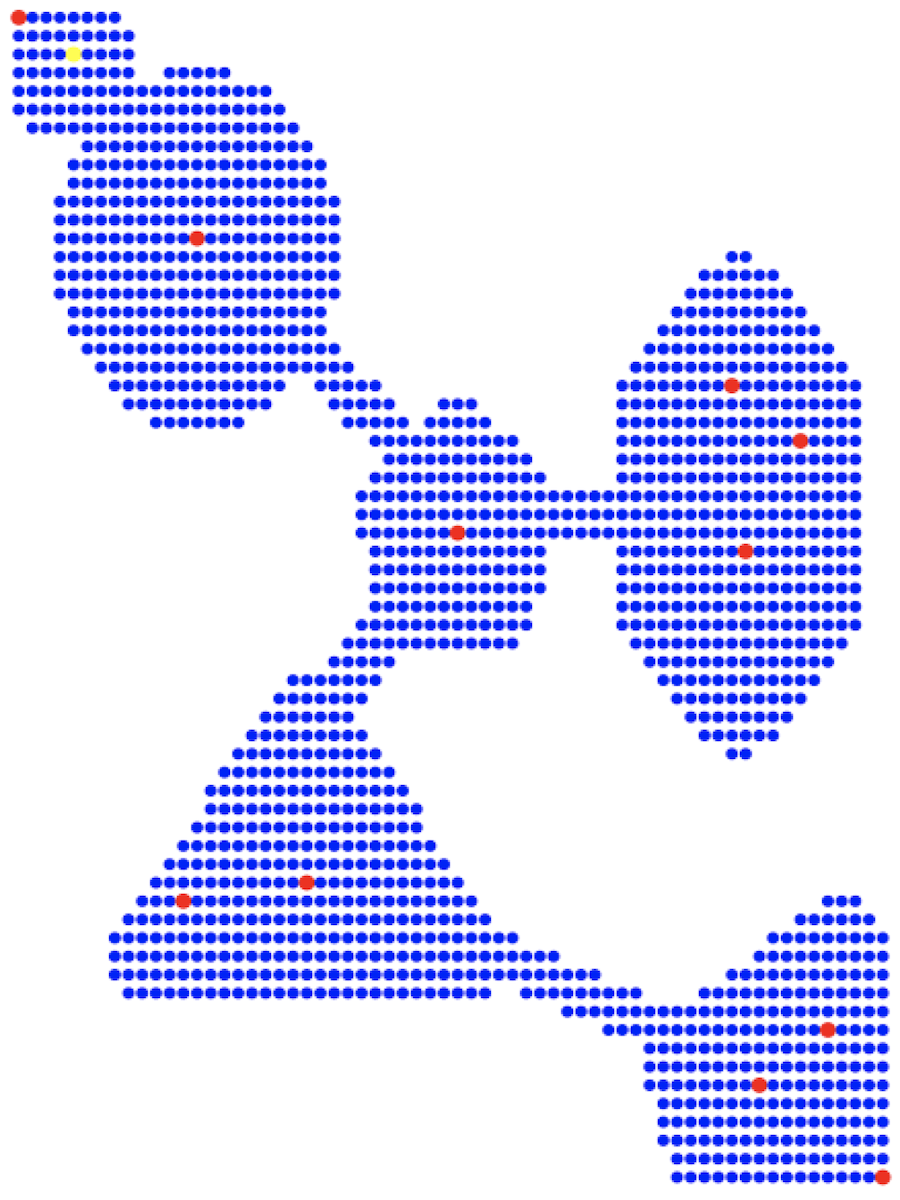}}}
   \subfloat[PI-$2$($158$, $458$)]{
    \label{fig:benchmark:3} 
    \fbox{\includegraphics[width=0.17\linewidth]{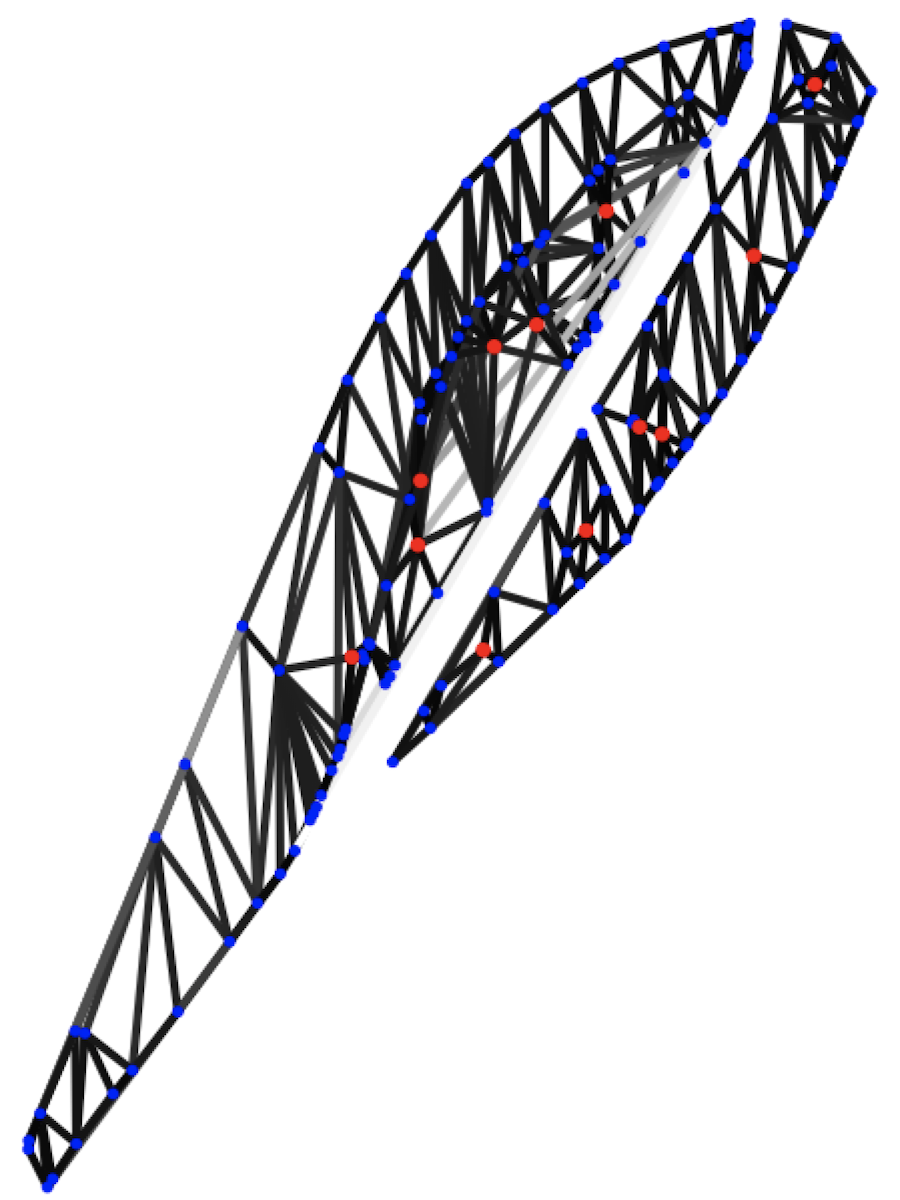}}}
   \subfloat[PI-$3$($888$, $2578$)]{
    \label{fig:benchmark:4} 
    \fbox{\includegraphics[width=0.17\linewidth]{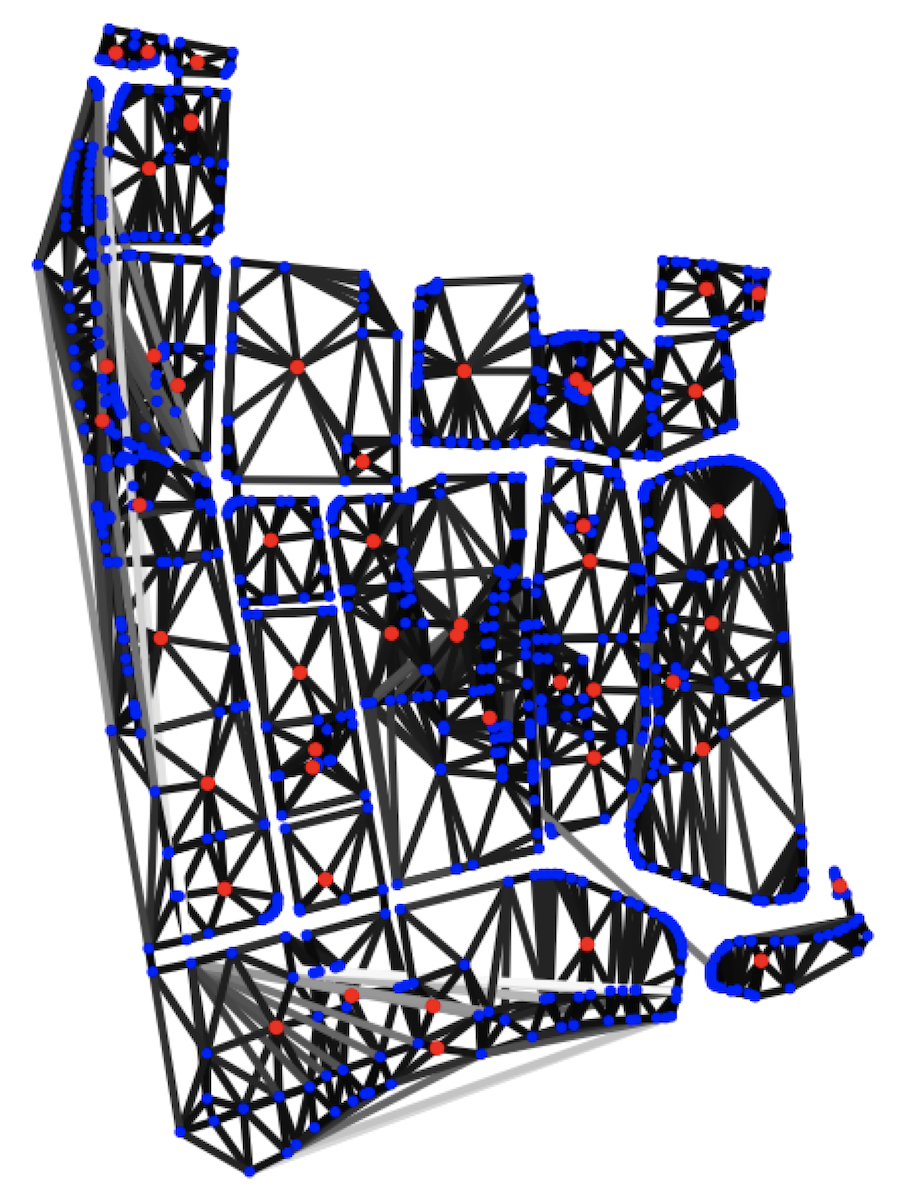}}}
   \subfloat[PI-$4$($1463$, $5268$)]{
    \label{fig:benchmark:2} 
    \fbox{\includegraphics[width=0.17\linewidth]{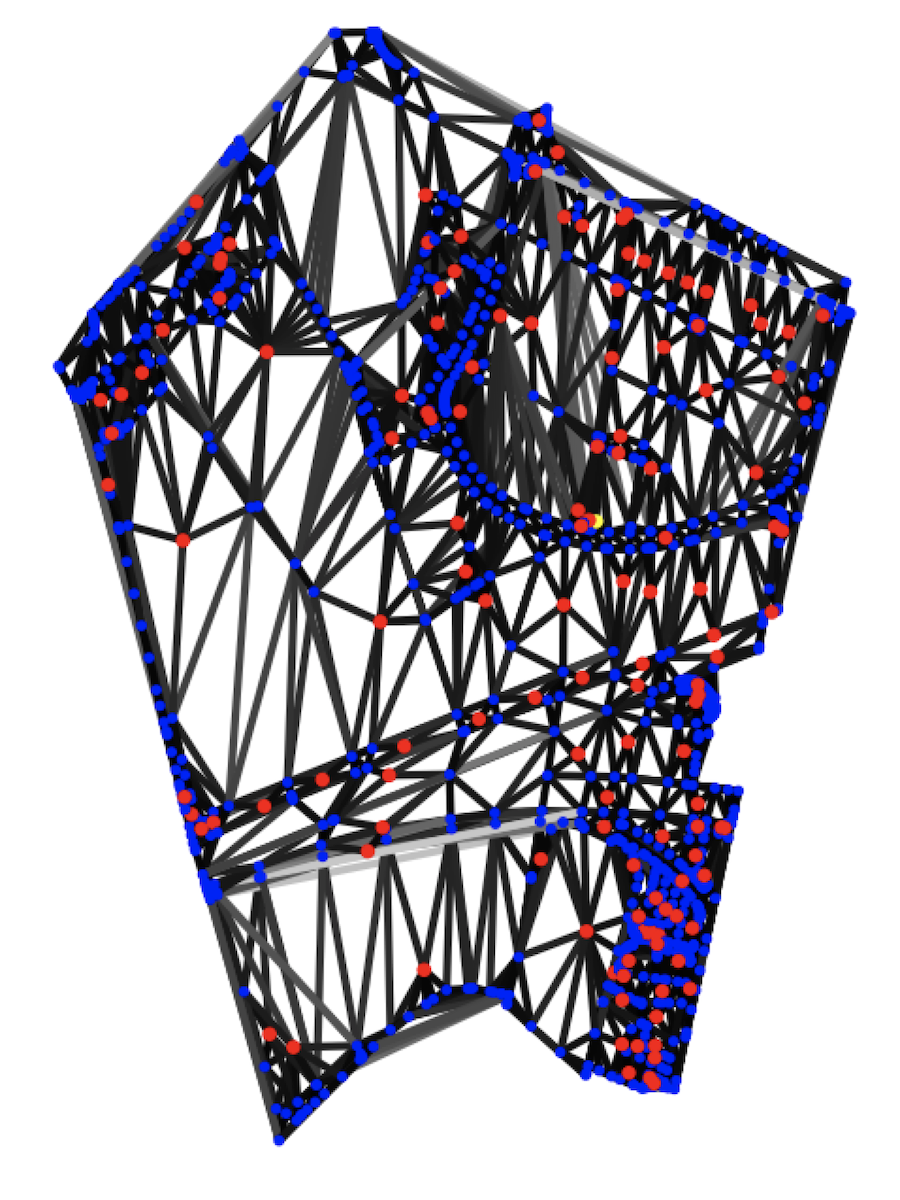}}}
   \subfloat[PI-$5$($3859$, $11163$)]{
    \label{fig:benchmark:5} 
    \fbox{\includegraphics[width=0.17\linewidth]{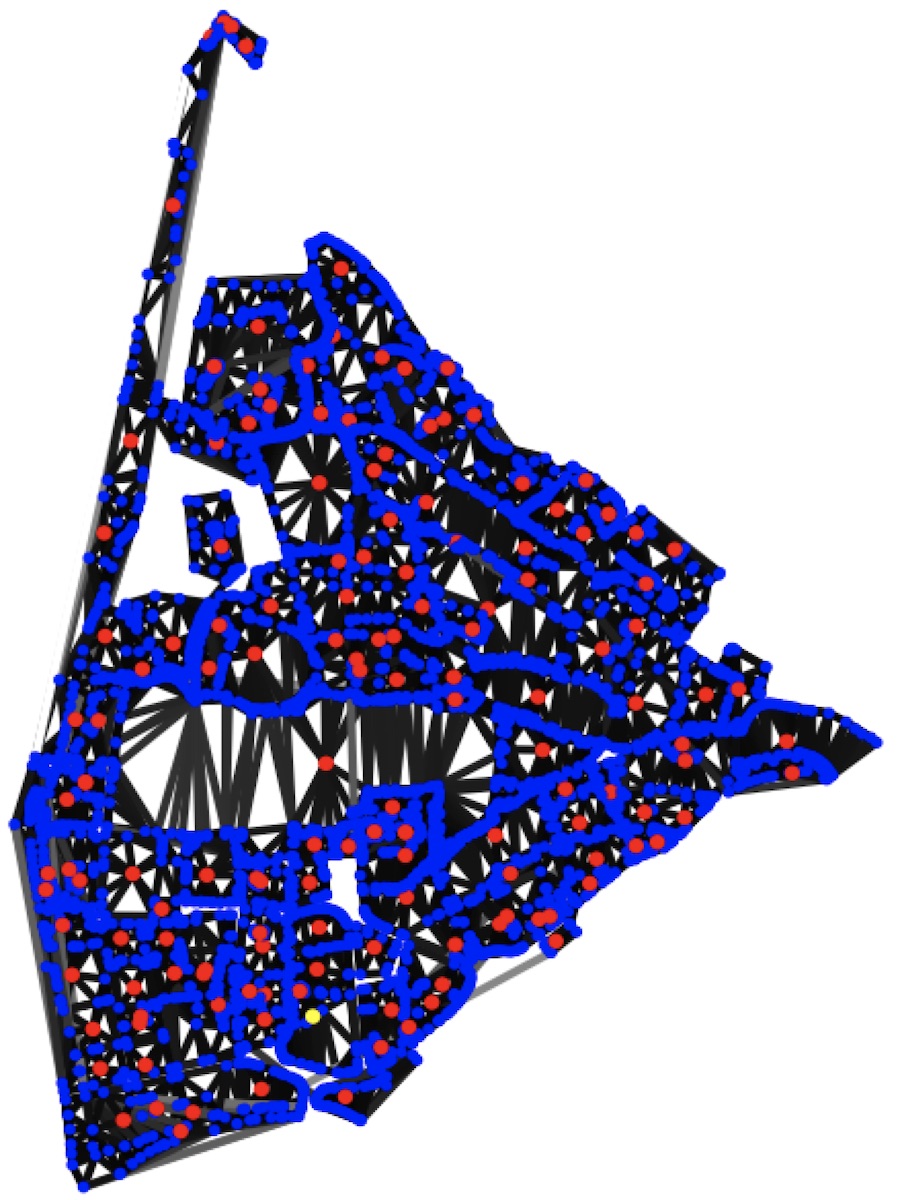}}}
 \caption[]{Benchmark problem instances denoted PI-$X$($N$, $E$) with varying numbers of nodes ($N$) and edges ($E$).}
 \label{fig:benchmark} 
\end{figure*}

\subsection{Benchmark Problems}
The benchmark problems consist of a manually created problem instance (PI-$1$) and four real-life problems with a varying number of edges and nodes (PI-$2$ - PI-$5$), see Figure \ref{fig:benchmark}. 
PI-$4$ is based on cadaster data from the municipality of Muenster, whereas the other non-synthetic problem instances are generated from geographical data of South Tyrol \footnote{http://geokatalog.buergernetz.bz.it/geokatalog}. 
Graph complexity differs for every problem (see Figure \ref{fig:benchmark}).

PI-$1$ comprises a large number of nodes and edges, making it the second hardest problem in terms of nodes. 
However, PI-$1$ has a clear structure, making it easier to partition. 
By this it should be shown that result quality is not only affected by the number of possible solutions but also by the node arrangement of the input graph.
In addition to the varying number of edges and nodes the input graphs of PI-$1$ and PI-$3$ show several clearly visible clusters with a low number of inter-cluster connections. 
In contrary, components in PI-$4$ and PI-$5$ show higher connectivity.

\subsection{Partitioning}
\begin{figure}[hpbt]
   \subfloat[Voronoi]{
   \label{fig:benchmark:3:v} 
   \fbox{\includegraphics[width=0.27\linewidth]{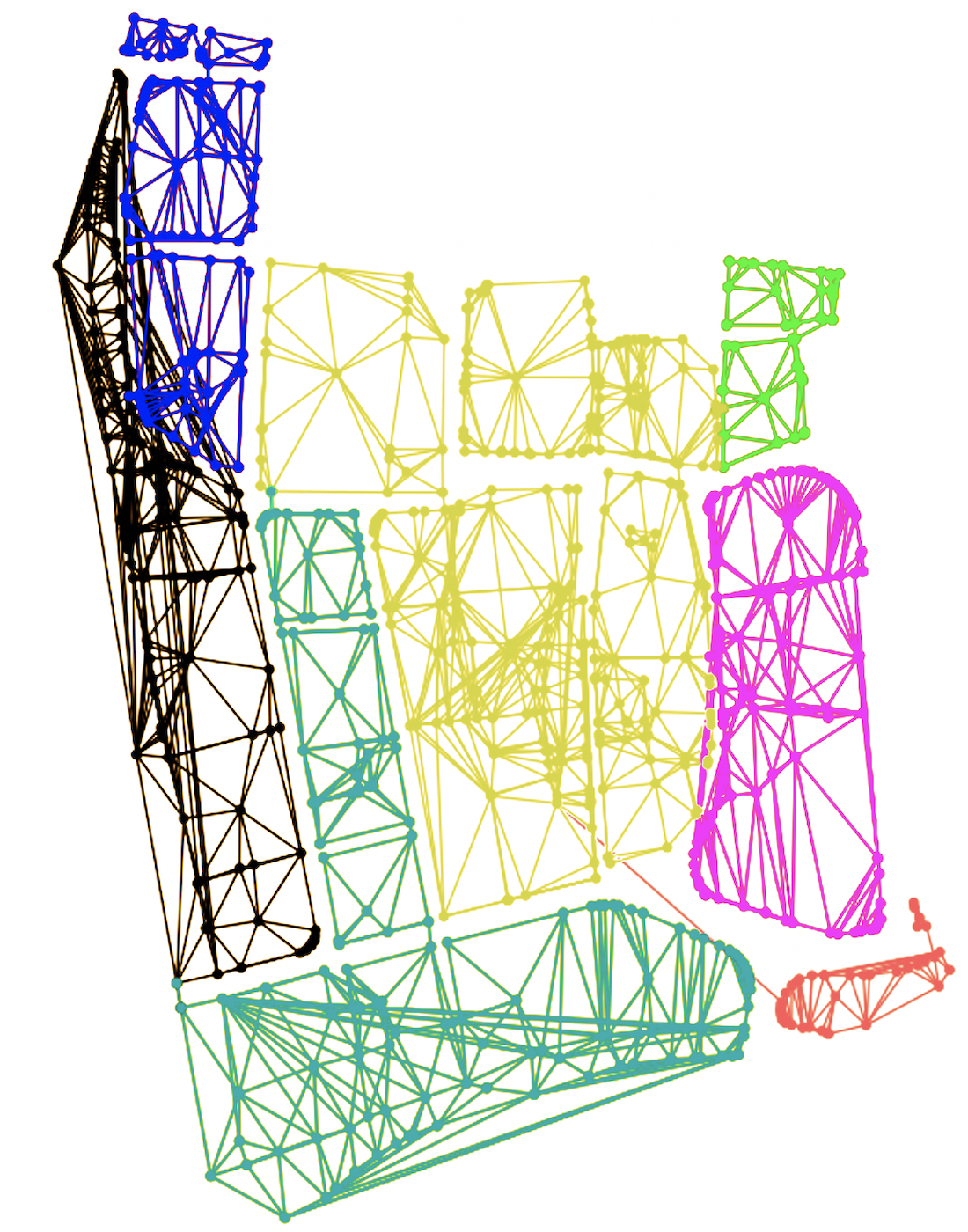}}}
   \subfloat[SC (EG)]{
   \label{fig:benchmark:4:sceg} 
   \fbox{\includegraphics[width=0.27\linewidth]{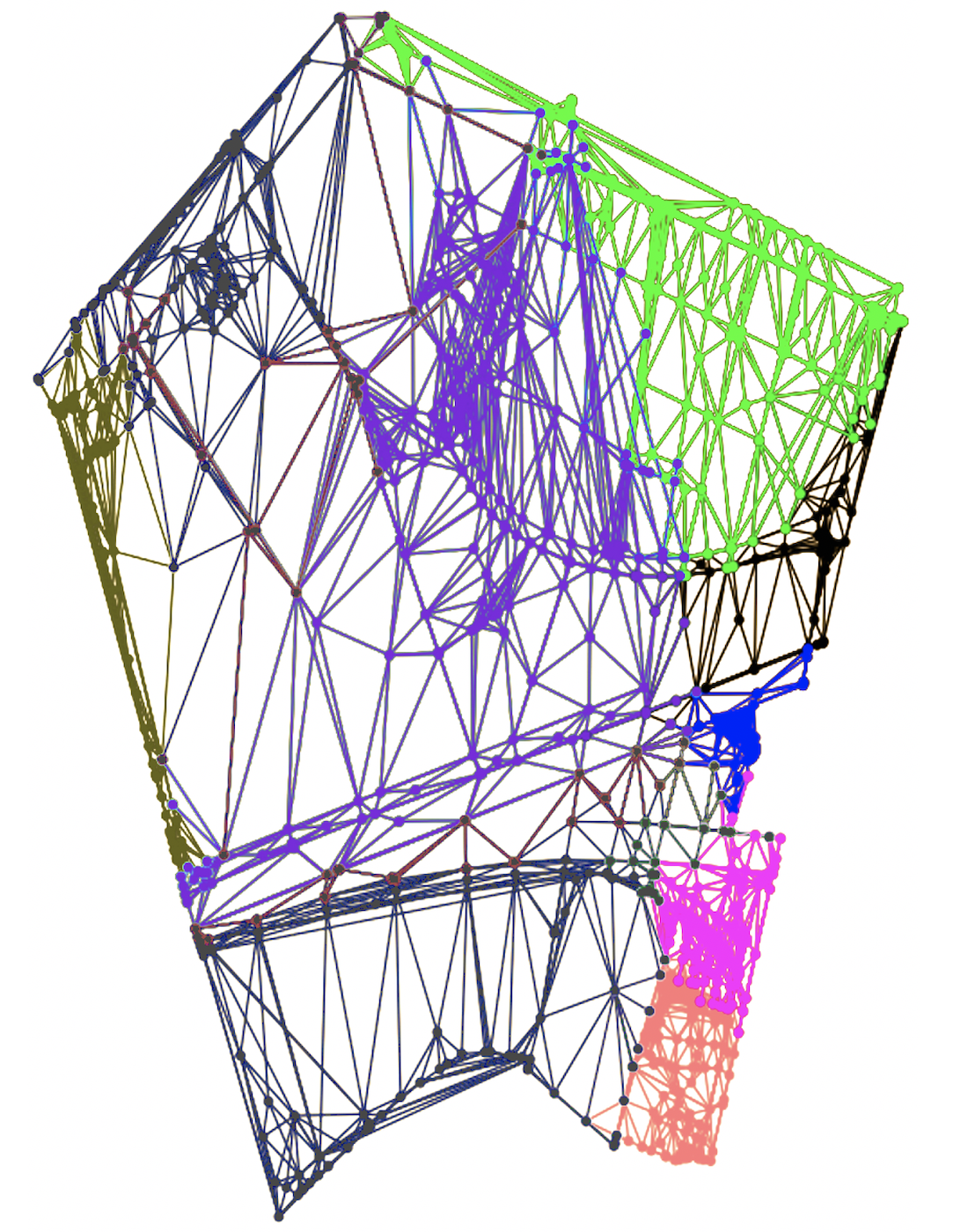}}}
   \subfloat[SC (V)]{
   \label{fig:benchmark:5:scv} 
   \fbox{\includegraphics[width=0.27\linewidth]{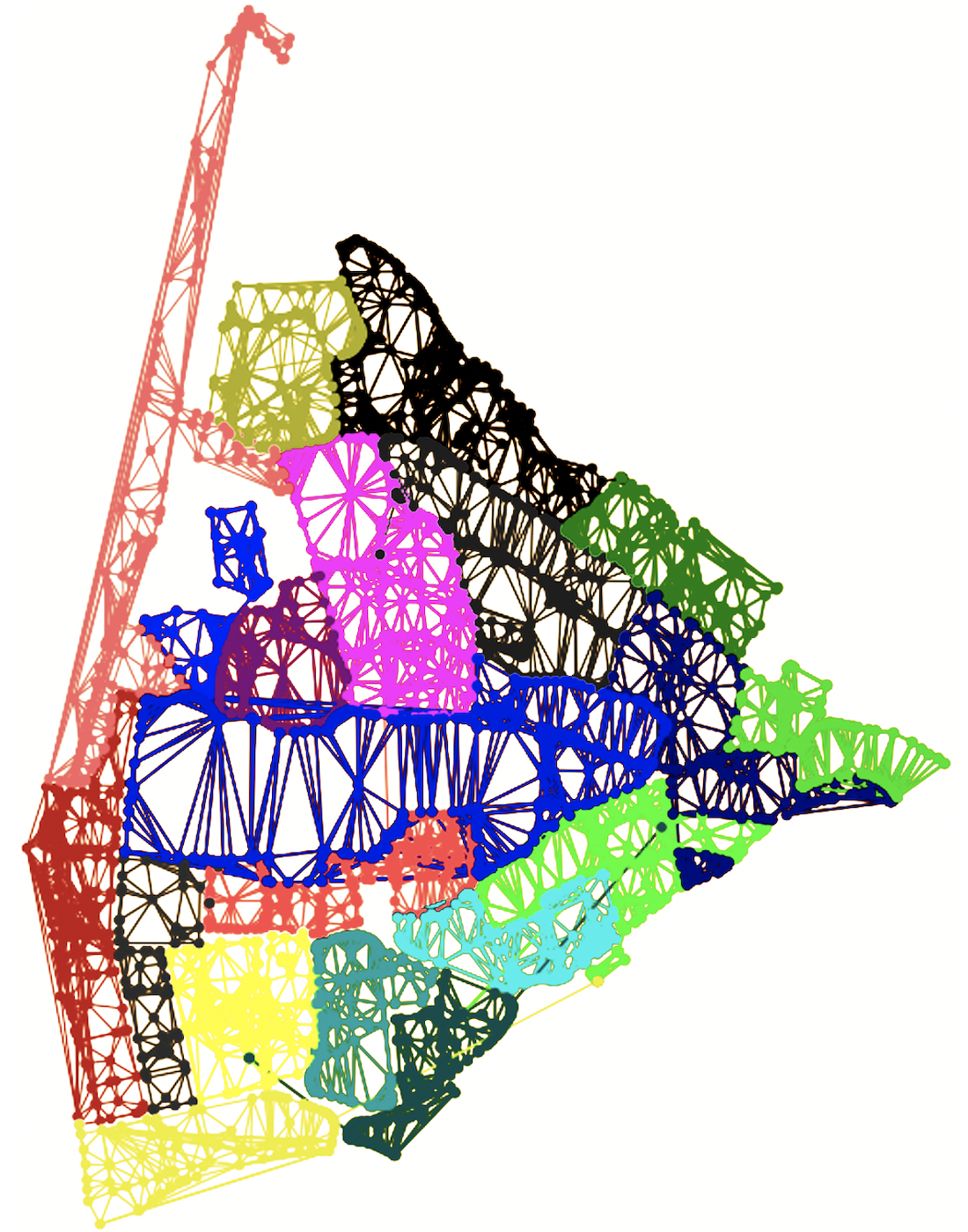}}}
 
 \caption[]{Comparison of partitioning methods: Voronoi (V), Spectral Clustering (with Eigen Gaps (EG) or Voronoi).}
 \label{fig:partitioner} 
\end{figure}

It is assumed that solvers produce more cost-efficient solutions of large Steiner Trees if large input graphs are partitioned into smaller subgraphs. 
Some well-partitioned input graphs can be seen in Figure \ref{fig:partitioner}, where different colors correspond to different subgraphs.

GM partitioning is well suited for smaller graphs with a low degree of connectivity between clusters (e.g. PI-$1$).

Voronoi-based partitioning produces well-separated subgraphs for bigger graphs with a low number of connections between subgraphs (see Figure \ref{fig:benchmark:3:v}). 
Spectral Clustering produces good clusters on all complexity levels. 
However, the predefined number of clusters based on Voronoi or Eigengaps seems to make a difference, especially for graphs with a strong meshing (e.g. PI-5).

To summarize, GM partitioning works best for input graphs with visible clusters and a low number inter-cluster connections.
If sparse graphs get larger, using partitioning Voronoi-based partitioning should produce better results. 
For dense graphs, Spectral Clustering is the best choice.

\subsection{Network Cost}

\begin{figure}[hpbt]
  \includegraphics[width=0.475\textwidth]{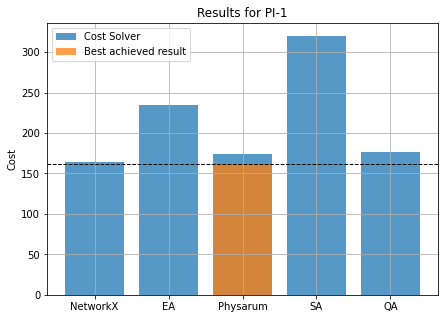}\\
  \caption{Network costs for PI-$1$. The blue bars show the best result achieved for each solver on the input graph without partitioning or simplification (except QA). As a reference, the orange bar shows the best overall result. This was achieved by a combination of Physarum and Greedy Modularity partitioning.}\label{fig:cost-pi-1}
\end{figure}

Figure \ref{fig:cost-pi-1} shows the lowest achieved network cost for each solver for PI-$1$ without partitioning or simplifying (except for Quantum Annealing, where prior simplifying is necessary due to hardware constraints).
For better comparison, the best overall result with enabled partitioning is also included.
The NetworkX solver outperforms all others. 
However, the Physarum and Quantum Annealing solvers show slightly worse, but competitive results. 
The best overall result was achieved by a combination of Physarum and Greedy Modularity partitioning. 
There, the baseline is outperformed by $1.22\%$. 
Interestingly, a partitioner or simplifier combined with the NetworkX approach results in increased cost. 

\begin{figure}[hpbt]
  \includegraphics[width=0.47\textwidth]{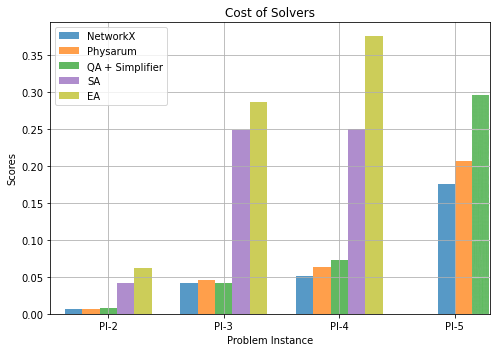}\\
  \caption{Resulting costs for PI-$2$ to PI-$5$ for each solver.}\label{fig:cost-solver}
\end{figure}

A similar correlation is visible for real-life problem instances. 
Figure \ref{fig:cost-solver} shows the results of each solver. 
Again, Quantum Annealing is the only algorithm with prior simplifying. 
NetworkX, Physarum and Quantum Annealing return the most cost-efficient solutions, outperforming the Evolutionary Algorithm and Simulated Annealing. 
Please note that PI-$5$ could only be solved by NetworkX and Physarum without the help of partitioners or simplifiers.

\begin{figure}[hpbt]
  \includegraphics[width=0.475\textwidth]{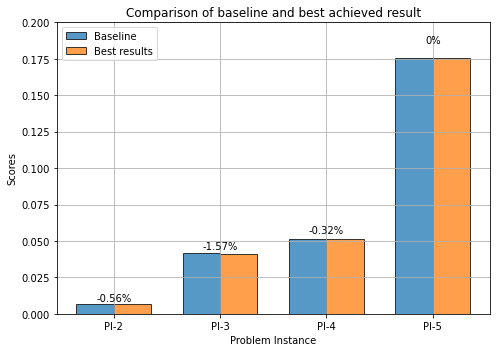}\\
  \caption{Comparison of baseline and best achieved results. The number above each problem denotes the cost improvement in percent. A negative value corresponds to a cost improvement.}\label{fig:best_results}
\end{figure}

Every possible combination of simplifier, partitioner and solver was evaluated on each instance. 
Figure \ref{fig:best_results} shows the best overall result compared with the NetworkX baseline. 
The number above each problem denotes the cost improvement in percent. 
Hence, the baseline could be outperformed on PI-$2$ by $0.56\%$, PI-$3$ by $1.57\%$ and PI-$4$ by $0.32\%$. 
Except for PI-$2$, these enhancements have been achieved with prior partitioning.

Table \ref{tab:solver-results} contains a more detailed overview of the lowest achieved costs for each solver and problem instance. 
Additionally, the combination of simplifier, partitioner and solver which returned the best result for each problem are highlighted. 
In summary, the NetworkX baseline could be outperformed on $4$ out of $5$ problem instances. 
Physarum in combination with prior partitioning is the overall most cost-efficient approach, where the choice of the partitioning scheme depends on the initial graph. 
Furthermore, simplification does not further reduce the lowest costs for any of the evaluated problem instances.

\begin{center}
\begin{table}[htbp]
{\small
\begin{center}
\begin{tabular}[center]{l p{3.7cm} c}
\toprule
Problem & Approach & Cost\\
\midrule
\multirow{3}{4em}{PI-1 \\ $N=1560$ \\ $E=4294$} 
& NetworkX & $164.0$\\ 
& EA & $235.0$\\ 
& Physarum ($5$ runs) & $174.0$\\ 
& Simulated Annealing & $320.0$\\
& QA (Leap) + GNG + SC (Voronoi) & $175.0$\\
& \textbf{Physarum + GM} & \textbf{$162.0$}\\
\hline
\multirow{3}{4em}{PI-$2$ \\ $N=158$ \\ $E=458$} 
& NetworkX & $0.006679$\\ 
& EA      & $0.062119$\\ 
& \textbf{Physarum ($5$ runs)} & \textbf{$0.006641$}\\ 
& Simulated Annealing & $0.042000$\\
& QA (qbsolv) + Triangle & $0.007861$\\
\hline
\multirow{3}{4em}{PI-$3$ \\ $N=888$ \\ $E=2578$} 
& NetworkX & $0.041419$\\ 
& EA      & $0.286315$\\ 
& Physarum ($5$ runs)                   & $0.045759$\\ 
& Simulated Annealing         & $0.249000$\\
& QA (Leap) + GNG & $0.042000$\\
& \textbf{Physarum + SC} & \textbf{$0.040768$}\\
\hline
\multirow{3}{4em}{PI-$4$ \\ $N=1463$ \\ $E=5268$} 
& NetworkX & $0.051680$\\ 
& EA      & $0.375550$\\ 
& \textbf{EA + NetworkX} & \textbf{$0.051514$}\\ 
& Physarum ($5$ runs)                    & $0.063590$\\ 
& Simulated Annealing & $0.250000$\\
& QA (qbsolv) + GNG + SC (Voronoi) & $0.072522$\\
\hline
\multirow{3}{4em}{PI-$5$ \\ $N=3859$ \\ $E=11163$} 
&  \textbf{NetworkX} & \textbf{$0.175814$}\\  
& EA & - \\ 
& Physarum ($5$ runs) & $0.206857$\\ 
& Simulated Annealing & -\\ 
& QA (qbsolv) + GNG + Voronoi & $0.280829$\\

\bottomrule
\end{tabular}
\end{center}
} 
\caption[Solver: Cost]{Network cost for different solvers: NetworkX baseline, Evolutionary Algorithm (EA), Physarum, Simulated Annealing, Quantum Annealing (QA). Other abbreviations: Growing Neural Gas (GNG), Spectral Clustering (SC), Greedy Modularity (GM), Triangulation (Triangle) }\label{tab:solver-results}
\end{table}
\end{center}

\subsection{Duration}
\begin{figure}[hpbt]
  \centering
  \includegraphics[width=0.47\textwidth]{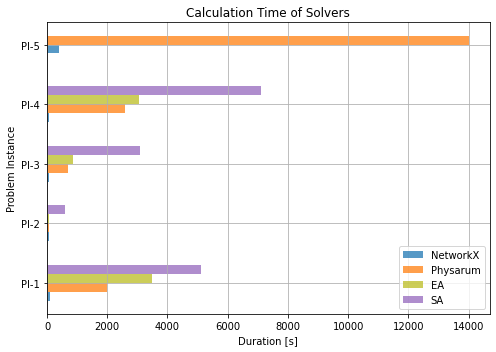}\\
  \caption{Computation times for all solvers and problem instances in [s].}\label{fig:duration}
\end{figure}

Figure \ref{fig:duration} shows the computation time for each solver and problem instance. 
Quantum Annealing was excluded from this graph since prior simplifying was needed and this would drastically affect computation time.
The NetworkX baseline is the fastest approach throughout all instances with small increases for each complexity level. 
The Evolutionary Algorithm and Physarum mostly show similar results, still outperforming Simulated Annealing. However, Physarum shows a large spike for PI-$5$, returning the result after almost $233$ minutes compared to $36$ minutes for PI-$4$. 
This effect could be reduced by partitioning but since the baseline algorithm could not be outperformed on this instance, no further experiments were conducted.

\section{\uppercase{Related Work}}
\label{sec:relatedwork}
Even though Steiner Tree Problems have already been studied for decades \cite{duin94,promel02}, it has been revisited repeatedly with a high variety of approaches.

For example, Rosenberg et al. \cite{rosenberg21} developed and successfully applied Evolutionary Algorithms to Euclidean STPs with soft obstacles and up to $1000$ nodes. Despite needing a relatively long time to converge, the approach was able to outperform an iterative approach in terms of cost-efficiency and showed promising results.

In \cite{siebert20} a technique based on Dynamic Programming and neighborhood structure characterization was proposed, which also uses Simulated Annealing. Recently, work on using gate-model Quantum Computers for solving the STP has been published \cite{miyamoto}. In their publication, Miyamoto et al. have devised a hybrid-quantum algorithm based on Grover's search algorithm \cite{grover96} that has a time complexity better than classical state-of-the-art algorithms. 
However, no experimental results are available and there exist no statements on necessary error correction strategies or on its applicability on Noisy Intermediate-Scale Quantum Computers (NISQ).
\section{\uppercase{Conclusion}}
\label{sec:conclusion}
In this work, a wide range of state-of-the-art algorithms for simplifying, partitioning and solving large STPs was introduced and discussed.
The proposed algorithms were evaluated on multiple real-life problem instances and compared against a widely used baseline algorithm which could be outperformed on $4$ out of $5$ instances. 
Even though good partitioning was key for achieving high network cost-efficiency, it is assumed that the partitioning and merging process affects result optimality negatively. 
Further work towards optimal partitioning of Steiner Trees might lead to better results. Also, input graph simplification did not lead to a further decrease in network costs.

Although computational time is comparatively long for Physarum, it is reasonable to leverage this approach since it outperformed the baseline by about $1\%$ on 3 out of 5 instances, which, due to high excavation costs, makes a significant contribution to cost savings when expanding FTTH networks.
Hence, longer runtime is mostly compensated by cost-efficiency, especially when approximating district-sized FTTH networks. Larger networks with a high degree of connectivity (PI-4 and PI-5) could not be outperformed by Physarum. However, the results on these instances could potentially be improved through better partitioning.

While Quantum Annealing hardware is not yet usable to solve STP instances with sizes relevant to our application domain, experiments have shown that running Quantum Annealing algorithms on simulators or hybrid software already returns decent results, especially for smaller problem instances.
This indicates that performing experiments on future Quantum Annealers with more physical qubits might still be of interest. There seems to be a trend for the number of qubits in Quantum Annealers to double every two to three years \cite{advantage}. In combination with a potentially more efficient QUBO formulation \cite{Fowler17}, it should thus be possible to solve STP instances for FTTH networks of relevant size within the next decade. With the help of sophisticated partitioners and simplifiers, this should be achievable even sooner.

\section*{\uppercase{Acknowledgements}}
The authors would like to thank Deutsche Telekom Laboratories (T-Labs) for funding this work.

\bibliographystyle{apalike}
{\small
\bibliography{main}
}

\end{document}